\documentclass{article}

\PassOptionsToPackage{numbers, compress}{natbib}

\usepackage[preprint]{neurips_2022}

\usepackage[utf8]{inputenc} 
\usepackage[T1]{fontenc}    
\usepackage{url}            
\usepackage{booktabs}       
\usepackage{amsfonts}       
\usepackage{nicefrac}       
\usepackage{microtype}      
\usepackage{xcolor}         

\usepackage{enumitem}
\usepackage{graphicx}
\usepackage{xspace}
\usepackage{bm}
\usepackage{amssymb}
\usepackage{enumitem}
\usepackage[pagebackref=true,breaklinks=true,letterpaper=true,colorlinks,urlcolor=black,bookmarks=false]{hyperref}
\hypersetup{citecolor=[RGB]{119,185,0}}

\newcommand{\eg}{\emph{e.g., }}

\newcommand{\ours}{MinVIS\xspace}

\newcommand{\tabfontsize}{}

\newcommand{\figref}[1]{Figure\,\ref{fig:#1}}
\newcommand{\secref}[1]{Section\,\ref{sec:#1}}
\newcommand{\tabref}[1]{Table~\ref{tab:#1}}

\title{
MinVIS: A Minimal Video Instance Segmentation Framework without Video-based Training
}

\author{
  De-An Huang \\
  NVIDIA\\
  \texttt{deahuang@nvidia.com} \\
  \And
  Zhiding Yu \\
  NVIDIA \\
  \texttt{zhidingy@nvidia.com} \\
  \And
  Anima Anandkumar \\
  Caltech, NVIDIA \\
  \texttt{anima@caltech.edu} \\
}

\begin{document}

\maketitle

\begin{abstract}
We propose \ours, a minimal video instance segmentation (VIS) framework that achieves state-of-the-art VIS performance with neither video-based architectures nor training procedures. By only training a query-based image instance segmentation model, \ours outperforms the previous best result on the challenging Occluded VIS dataset by over 10\% AP. Since \ours treats frames in training videos as independent images, we can drastically sub-sample the annotated frames in training videos without any modifications. With only 1\% of labeled frames, \ours outperforms or is comparable to fully-supervised state-of-the-art approaches on YouTube-VIS 2019/2021. Our key observation is that queries trained to be discriminative between intra-frame object instances are temporally consistent and can be used to track instances without any manually designed heuristics. \ours thus has the following inference pipeline: we first apply the trained query-based image instance segmentation to video frames independently. The segmented instances are then tracked by bipartite matching of the corresponding queries. This inference is done in an online fashion and does not need to process the whole video at once. \ours thus has the practical advantages of reducing both the labeling costs and the memory requirements, while not sacrificing the VIS performance. Code is available at: \url{https://github.com/NVlabs/MinVIS}
\end{abstract}

\section{Introduction}

Video instance segmentation (VIS) aims to simultaneously detect, segment, and track object instances in videos~\cite{yang2019video}. The requirement to accurately track object instances through an entire video makes VIS much more challenging than image instance segmentation. Most of the early approaches for VIS build on image instance segmentation models, and process videos on a \emph{per-frame} basis~\cite{yang2019video,yang2021crossover}. The segmented object instances for each frame are then matched temporally with a post-processing step. This post-processing step often involves manually designed heuristics that do not generalize well to challenging scenarios like occlusions and large appearance deformations.

Recent VIS works address this issue by taking a \emph{per-clip} approach, where the spatial-temporal volume of a video is processed as a whole to directly predict the spatial-temporal mask for each object instance~\cite{cheng2021mask2former,wang2021end,wu2021seqformer}. Many of these end-to-end VIS approaches are built upon the recent advances of Transformers for end-to-end object detection~\cite{carion2020end}. 
Given learned embeddings called \emph{queries}, Transformers process the queries jointly with the input video using cross-attention, so that each of the processed queries can be used to predict the spatial-temporal mask for an object instance in the video.

While these per-clip methods have led to considerable improvements for VIS, using attention to process the whole video, especially longer ones,  requires large memory  and computation. It is also not straightforward to adapt per-clip methods from offline to online processing to reduce the computational requirements. This limits their practical application, and maintaining the effectiveness of these per-clip methods while improving their efficiency remains an active research direction~\cite{hwang2021video,yang2022tevit}.

Another limitation for existing VIS methods is the requirement on annotation. Annotating object instance masks for each video frame is prohibitively expensive at scale. While there have been works that alleviate this annotation requirement through weak supervision or image-based annotation, there is still a significant performance gap compared to state-of-the-art fully-supervised methods~\cite{fu2021learning,liu2021weakly}.

\noindent\textbf{Our Approach.} We simultaneously address both of the aforementioned challenges of computational and labeling costs by showing that we can achieve state-of-the-art VIS performance by only training a query-based \emph{image} instance segmentation model. 
During inference, \ours first applies the query-based image instance segmentation to video frames independently. The segmented instances are then associated by bipartite matching of the corresponding queries.
\ours processes each frame independently in an online fashion and does not need to process the whole video at once. \ours does not use any video-based training procedure and thus does not need annotations for all the frames in a video. 
Our contributions are summarized below:
\vspace{-0.05in}
\begin{enumerate}[leftmargin=*,itemsep=0mm]
    \item We show that video-based architecture and training are not required for competitive VIS performances. \ours outperforms previous state-of-the-art on YouTube-VIS 2019 and 2021 datasets by 1\% and 3\% AP while only training an \emph{image} instance segmentation model.
    \item We show that image instance segmentation models capable of segmenting occluded instances are also well suited to track occluded instances in videos in our framework. 
    \ours outperforms its per-clip counterpart by over 13\% AP on the challenging Occluded VIS (OVIS) dataset, which is over 10\% improvement compared to the previous best performance on the dataset.
    \item Our image-based approach allows us to significantly sub-sample the required segmentation annotations in training without any change to the model. With only 1\% of labeled frames, \ours outperforms or is comparable to fully-supervised state-of-the-art approaches on all three datasets.
\end{enumerate}

\begin{figure}
  \centering
  \includegraphics[width=0.95\linewidth]{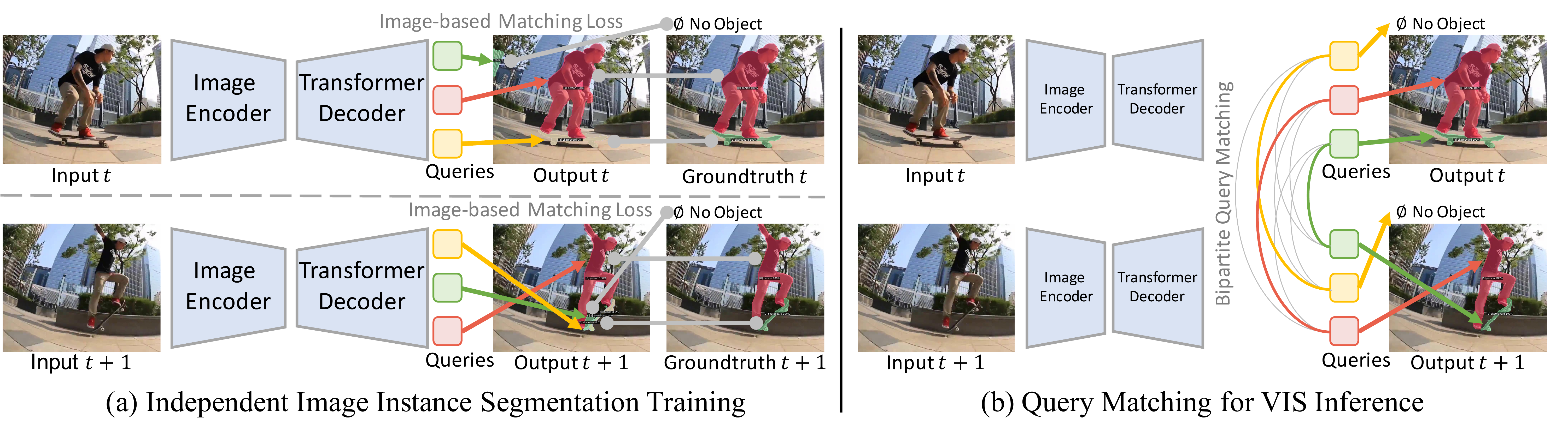}
  \caption{(a) \ours trains a query-based image instance segmentation model (Image Encoder + Transformer Decoder) using each frame independently. (b) During inference, the trained image instance segmentation model is used for video instance segmentation by bipartite matching of query embeddings across frames. \ours does not require further manually designed heuristics for tracking.}
  \label{fig:fig1}
\end{figure}

Our key observation is that queries trained to be discriminative between intra-frame object instances are temporally consistent and can be used to track instances without being trained with video-based loss functions.
\ours achieves this by requiring its image instance segmentation model to generate masks by convolving query embeddings with features of the whole input image, including regions of other object instances.
A query is thus trained to only have high responses on features of its corresponding instance. Other query embeddings should instead have low responses on these features because instance masks are non-overlapping.
This design encourages the query embeddings for different instances in a frame to be well-separated.
On the other hand, the query embeddings that segment the same instance from two consecutive frames still need to be similar enough since the instance's image features to be convoluted do not change drastically between frames.
This leads to temporally consistent query embeddings for tracking without the need of video-based training.

\ours thus has the following design for inference: We first apply a query-based image instance segmentation model on video frames independently. The segmented instances are then associated between frames by bipartite matching of the corresponding query embeddings. This post-processing step does not need any additional heuristics based on mask overlaps or classification scores as in previous works~\cite{yang2019video,QueryTrack}. This is because query embeddings already contain these information to track the instances.
An overview of \ours's training and inference is shown in \figref{fig1}.

Since video frames are treated as independent images to train \ours, there is no requirement to annotate all the frames in a video for training. This allows us to significantly sub-sample and reduce the annotation without any change to our model. We find that on YouTube-VIS 2019/2021 datasets~\cite{yang2019video}, where there are less variations between video frames, using only 1\% of labeled frames leads to less than 3\% drop in AP for \ours.

We further evaluate \ours on the Occluded VIS (OVIS) dataset~\cite{qi2021occluded}. One common critique of per-frame approaches is that their tracking heuristics based on mask overlaps would not work when there are heavy occlusions. This is not the case for \ours, as we do not use any manually designed heuristics. We show that our query-matching approach generalizes to occluded scenarios. \ours with Swin Transformers backbone~\cite{liu2021swin} achieves 39.4\% AP on OVIS, which is over 10\% improvement from the previous best result on the dataset~\cite{li2021limited}. We further show that our image-based strategy leads to easier and better learning on OVIS. \ours outperforms its per-clip counterpart by over 13\% AP.

\section{Related Work}

\noindent\textbf{Video Instance Segmentation.} 
Per-frame approaches for VIS process each frame independently and later track instances by post-processing. MaskTrack R-CNN~\cite{yang2019video} adds a tracking head to Mask R-CNN~\cite{he2017mask} for VIS.
MaskProp~\cite{bertasius2020classifying} instead adds a mask propagation head to propagate object instance masks. CrossVIS~\cite{yang2021crossover} uses crossover learning  to improve instance representation across video frames. QueryTrack~\cite{QueryTrack} adds a contrastive tracking head to QueryInst~\cite{Fang_2021_ICCV} for instance association. Concurrent work IDOL~\cite{IDOL} shows that per-frame models can still outperform per-clip models.
Our approach also builds on image instance segmentation models, but unlike previous approaches, we need neither additional parameters nor additional losses to apply to VIS. Our query embeddings from image instance segmentation can directly be used for tracking without video-based training.

Recent per-clip approaches build on the success of Detection Transformer (DETR)~\cite{carion2020end}. VisTR~\cite{wang2021end} adopts the query-based approach of DETR to VIS, and there has been several follow-up works, such as Mask2Former-VIS~\cite{cheng2021mask2former} and SeqFormer~\cite{wu2021seqformer}. One limitation of these approaches is the need to process the whole video at once.
IFC~\cite{hwang2021video} reduces the overhead of temporal message passing by using memory tokens. TeViT~\cite{yang2022tevit} uses a parameter-shared self attention to efficiently model temporal contexts. We also use a query-based approach, but instead of using cross-attention to process the whole video, we process each frame independently while not losing VIS performance. 
Our use of queries to associate instances is also related to other works that build on DETR for tracking in related fields. For example, MOTR~\cite{zeng2021motr} and TrackFormer~\cite{meinhardt2021trackformer} use identity preserving track queries for multi-object tracking (MOT).

\noindent\textbf{Reducing Supervision for Video Instance Segmentation.} Annotating instance masks for each video frame can be prohibitively expensive. Compared to video object segmentation~\cite{lu2020learning,voigtlaender2021reducing,yang2021dystab} and image instance segmentation~\cite{ahn2019weakly,lan2021discobox,tian2021boxinst,wang2022freesolo}, there have been less works on reducing supervision for VIS~\cite{wang2021survey}. FlowIRN~\cite{liu2021weakly} extends IRN~\cite{ahn2019weakly} with motion and temporal consistency cues to have a weakly-supervised VIS framework that only requires classification labels. SOLO-Track~\cite{fu2021learning} learns to track instances without video annotations. It uses instance contrastive learning on SOLO~\cite{wang2020solo} to learn grid cell embeddings for instance tracking. We make the same observation that disciminating between instances within frames is beneficial or even sufficient for instance tracking. However, unlike our query-based association, the grid cell embeddings still need threshold-based post-processing and additional loss functions to better handle birth and death of objects.

\section{Method}

\ours is a minimal VIS framework that does not require video-based training and thus can be easily applied to real-world applications that only have sparse image instance segmentation annotations. \ours is a two stage approach: (1) image instance segmentation on each frame independently, (2) associating instances between frames by matching queries.
We will first discuss the image instance segmentation architecture in \ours. We will then discuss the temporal association of object instances. Finally, we will discuss training and reducing supervision for \ours. 

\begin{figure}
  \centering
  \includegraphics[width=0.95\linewidth]{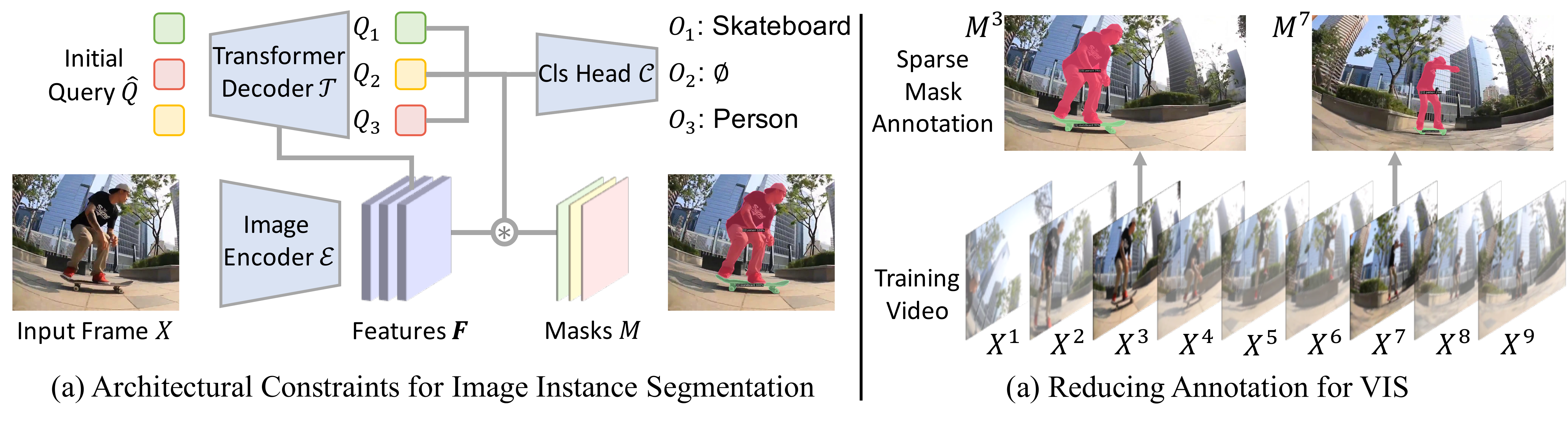}
  \caption{(a) \ours's main architectural constraint is to require the segmentation masks $M$ be generated by convolving the query embeddings $Q$ with the final feature map $F_{-1}$. This makes the query embeddings discriminative between each instances. (b) \ours's image-based approach allows direct annotation subsampling of training videos without any modification to the model.}
  \label{fig:system}
\end{figure}

\subsection{Image Instance Segmentation Architecture for VIS}
\label{sec:arch}

\ours builds on the query-based transformer architectures for detection and segmentation~\cite{carion2020end,Fang_2021_ICCV,cheng2021masked,zhu2021deformable}, which has the following main components: (1) \emph{Image Encoder} that learn to extract features from input images. (2) \emph{Transformer Decoder} that processes the outputs of Image Encoder to iteratively update the query embeddings. (3) \emph{Prediction Heads} that use the final query embeddings to predict desired outputs (\eg segmentation masks, class labels). The queries play an important role for the success of such end-to-end pipeline for set prediction with unknown number of outputs. The number of queries are selected as the maximum number of output instanes of the model. During inference, a subset of queries predict $\varnothing$ outputs to dynamically adjust the number of valid outputs.

An overview of \ours's image instance segmentation is shown in \figref{system}(a). Given an image $X \in \mathbb{R}^{H, W}$, the Image Encoder $\mathcal{E}$ extracts a set of features $\bm{F} = \mathcal{E}(X)$ from the image. $\bm{F} = \{F_0 \dots F_{-1}\}$ is a sequence of multi-scale feature maps $F_i \in \mathbb{R}^{H_i, W_i, C_i}$. $F_{-1}$ denotes the final output of $\mathcal{E}$. The $N$ initial query embeddings $\hat{Q} \in \mathbb{R}^{N,C}$ are learnable parameters, where $N$ is a large enough number of outputs. The Transformer Decoder $\mathcal{T}$ then take both $\bm{F}$ and $\hat{Q}$ to iteratively obtain $Q = \mathcal{T}(\bm{F}, \hat{Q}), Q\in \mathbb{R}^{N,C}$. While most recent works focus on the design of $\mathcal{T}$ to better process $\bm{F}$ for $Q$, \ours's architectural constraints are on the Prediction Heads. There are two outputs for each instance: classification and segmentation mask. The classification scores $O=\mathcal{C}(Q), O \in \mathbb{R}^{N,K}$ for $K$ classes are the output of Classification Head $\mathcal{C}$, and $Q$ should contain the class information for each instance. For segmentation masks $M \in \mathbb{R}^{N,H,W}$, \ours requires that $M$ be generated by convolving the query embeddings $Q$ with the final feature map $F_{-1}$. The shape for $F_{-1}$ is thus $H,W,C$. We have $M = \sigma(Q * F_{-1}$), where $\sigma(\cdot)$ is the sigmoid function. 

\begin{figure}
  \centering
  \includegraphics[width=0.94\linewidth]{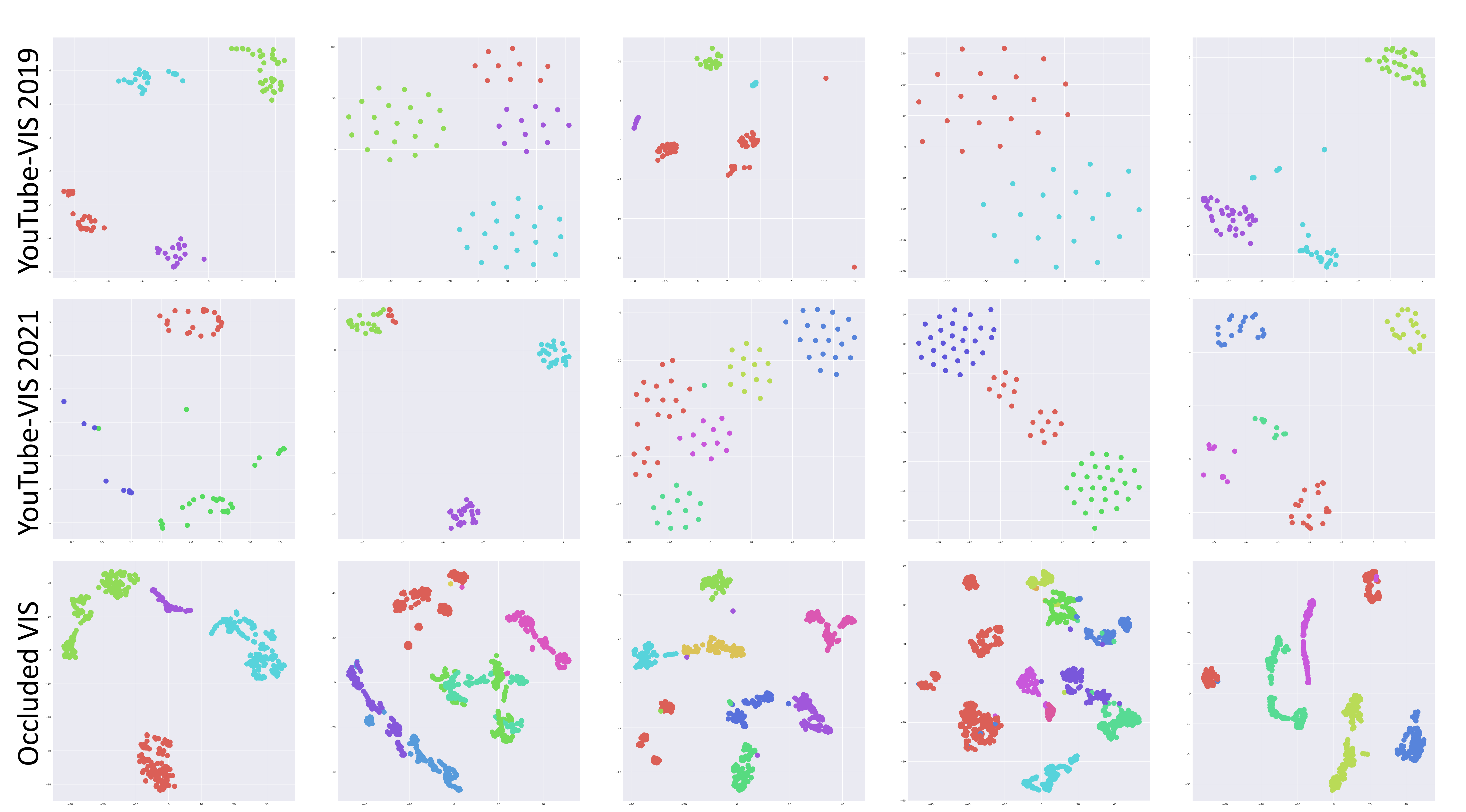}
  \caption{Visualizing our learned query embeddings with only image-based training. Each plot is for a video, and query embeddings of the same instance (from different frames) have the same color. Query embeddings are already grouped into clusters by instance without any video-based training.}
  \label{fig:tsne}
\end{figure}

The constraint to have $Q$ convolve with the whole feature map $F_{-1}$ is important for \ours. Consider two queries $Q_i$ and $Q_j$ that corresponds to two distinct object instances and thus non-overlapping masks. This formulation ensures that $Q_i$ should only have high inner products with features in $F_{-1}$ that are covered by the mask of instance $i$. Since the instance masks are non-overlapping, $Q_j$ should instead have low inner products with these features. This implicitly constrains the query embeddings to be discriminative between each other. 
On the other hand, if we apply this pipeline to two consecutive frames $X^t$ and $X^{t+1}$. Then $Q_i^{t}$ should still have higher inner product with $Q_i^{t+1}$ compared to $Q_j^{t+1}$. This is because $Q_i^{t+1}$ and $Q_j^{t+1}$ are also discriminative between each other, while $Q_i^{t}$ and $Q_i^{t+1}$ both need to have high inner products with features of the same instance, which do not change drastically between consecutive frames. We visualize our learned query embeddings in \figref{tsne}. Each plot is for a video. Query embeddings belonging to the same instance (from different frames) have the same color. These embeddings are already grouped by instances without any video-based training. Further details are in \secref{analysis}.

While not all image instance segmentation models satisfy our architectural constraints (\eg ROI-based architectures), we believe these are rather flexible designs that are compatible with various query-based instance segmentation models. We use Mask2Former~\cite{cheng2021masked} in this work. The Image Encoder $\mathcal{E}$ includes both the backbone and the pixel decoder of Mask2Former. We also find that having fully-connected layers to further process $Q$ before convolution is beneficial to the performance.

\subsection{Tracking by Query Matching}
\label{sec:match}

\ours is a per-frame two-stage approach, which requires a post-processing step to temporally associate instances. This post-processing often involves heuristics like mask overlaps, which does not generalize well to scenarios with heavy occlusions. 
Unlike previous two-stage approaches, \ours associate instances solely based on the query embeddings $Q$. 
Given two consecutive frames $X^t$ and $X^{t+1}$. We have $Q^t = \mathcal{T}(\bm{F}^t, \hat{Q})$, where $\bm{F}^t = \mathcal{E}(X^t)$, and similarly for $Q^{t+1}$. $Q^t_i$ is the query embedding for instance $i$. Tracking in \ours is done by the assignment of applying the Hungarian algorithm on a score matrix $S$, where $S_{ij} = cos(Q^t_i, Q^{t+1}_j)$. $cos(\cdot, \cdot)$ is the cosine similarity.

This approach is less affected by occlusions because each instance is represented by a query that does not have a spatial extent. 
In addition, we do not need heuristics to handle the birth and death of object instances in this framework. Since the number of queries is larger than the number of instances, there are queries that produce empty masks. The death of an object instance happens when its embedding is matched to such a null query. On the other hand, the birth of an instance is correctly handled if the matched query embeddings have been null before the actual birth of the object instance.

\subsection{Training with Less Supervision for VIS}
\label{sec:training}

Since the matching process does not need training, only the image instance segmentation model needs to be trained. There are two outputs of the model: classification scores $O \in \mathbb{R}^{N,K}$ and segmentation masks $M \in \mathbb{R}^{N,H,W}$ for $N$ queries, $K$ object classes, and image size $H$, $W$. We can process the groundtruth video instances to groundtruth image instances $O^* \in \mathbb{R}^{L,K}$ and $M^* \in \mathbb{R}^{L,H,W}$, where $L$ is the number of groundtruth instances ($N >> L$) and $O^*_j$ is a one-hot vector of groundtruth class. Given a loss function $\mathcal{L}(O_i, M_i, O^*_j, M^*_j)$ between predicted instance $i$ and groundtruth instance $j$, we first use bipartite matching to find the assignments between predicted and groundtruth instances that minimize the overall loss function, and train on those matched predictions with the loss function. 

More specifically, there are two terms in the loss function: $\mathcal{L}_{cls}$ and $\mathcal{L}_{mask}$. We use cross entropy loss for $\mathcal{L}_{cls}$ and binary cross entropy plus dice loss~\cite{milletari2016v} for $\mathcal{L}_{mask}$ as in previous works~\cite{cheng2021masked}. Both terms are purely image-based. The groudtruth video instances are first processed to instances for each frame independently. Therefore, even if there are only sparse frames labeled with instance, we can still train our model with the annotated frames. This provide a straightforward way to reduce the supervision for VIS. \figref{system}(b) shows the annotation sub-sampling to reduce supervision.

\section{Experiments}
\label{sec:exp}

\noindent\textbf{Datasets.} We evaluate \ours on three datasets: YouTube-VIS 2019/2021~\cite{yang2019video} and Occluded VIS (OVIS)~\cite{qi2021occluded}. The YouTube-VIS datasets contain 40 object classes. YouTube-VIS 2019 contains 2238/302/343 videos for training/validation/testing, while YouTube-VIS 2021 expands the dataset to 2985/421/453 videos for training/validation/testing, and includes higher quality annotations. OVIS has 25 object classes and contains 607/140/154 for training/validation/testing. While the number of videos is smaller, OVIS has more objects per frame, and the videos are also longer. This leads to more annotated instance masks compared to the YouTube-VIS datasets. In addition, OVIS also has much higher Bounding-box Occlusion Rate (0.22 v.s. 0.06/0.07) compared to the YouTube-VIS datasets, which indicates heavier occlusions between object instances.

\noindent\textbf{Metrics.} We follow previous works and use Average Precision (AP) and Average Recall (AR) as evaluation metrics~\cite{yang2019video}. AP is computed based on 10 intersection-over-union (IoU) thresholds from 50\% to 95\% with 5\% increment. The reported AP and AR are first computed for each object class and then averaged over all classes. All three datasets have public evaluation servers.

\noindent\textbf{Baselines.} We focus on results using ResNet50 and Swin-L backbones. ResNet50 is still the most widely used backbone for VIS, while Swin-L gives the best performances. Not all methods report both backbones on all three datasets. We include results that are available. For YouTube-VIS datasets, we include recent state-of-the-art results from SeqFormer~\cite{wu2021seqformer}, TeViT~\cite{yang2022tevit}, and Mask2Former-VIS~\cite{cheng2021mask2former}. These are all Transformer-based per-clip approaches as this paradigm has been recently dominating the field. On the other, out of of these methods, only TeViT is applied to OVIS. Therefore, we further compare to CMaskTrack R-CNN~\cite{qi2021occluded}, CrossVIS~\cite{yang2021crossover}, and STC~\cite{jiang2022stc}. These are all methods that allow online processing. Even TeViT uses a near online inference for OVIS~\cite{athar2020stem}. This is because OVIS has longer videos that would lead to out-of-memory for most of the per-clip approaches. 

Out of all the baselines,  Mask2Former-VIS~\cite{cheng2021mask2former} is the most related to \ours, as \ours is built on Mask2Former in this work. Mask2Former-VIS thus can be seen as the per-clip version of ours and is an important baseline for comparison. Therefore, we further apply Mask2Former-VIS on the OVIS dataset. Due to memory constraints, the videos in OVIS are first split into clips of length 30. We use the same post-processing as \ours to merge the outputs from these clips.

\noindent\textbf{Implementation Details.} 
Unless otherwise noted, our hyper-parameters follow Mask2Former-VIS~\cite{cheng2021mask2former}. All models are pre-trained with COCO instance segmentation~\cite{lin2014microsoft}. For OVIS, we use the same hyper-parameters as YouTube-VIS 2019 except training for $10k$ iterations instead of $6k$. For training losses, the weights are $5.0$ for $\mathcal{L}_{mask}$ and $2.0$ for $\mathcal{L}_{cls}$. All results of \ours are averaged over 3 random seeds. We sub-sample training to X\% by uniformly sampling frames in the video. We set a minimum of 1 frame per video. Since YouTube-VIS datasets often have videos less than a hundred frames. Our 1\% results are better seen as one frame per video results for YouTube-VIS.

\subsection{Main Results}

\begin{table}
  \caption{YouTube-VIS 2019 results. C80k indicates joint training with COCO images that have YouTube-VIS categories. \ours with X\% means sub-sampling the annotated frames in training.}
  \label{tab:ytvis2019}
  \centering
  \tabfontsize
  \begin{tabular}{lllccccc}
    \toprule
Method         & Backbone & Training  & AP   & AP$_{50}$ & AP$_{75}$ & AR$_1$ & AR$_{10}$ \\\midrule
TeViT~\cite{yang2022tevit}           & R50  &Full      & 42.1 & 67.8    & 44.8    & 41.3   & 49.4    \\
TeViT~\cite{yang2022tevit}           & MsgShifT  &Full      & 46.6 & \textbf{71.3}    & 51.6    & 44.9   & 54.3    \\
SeqFormer~\cite{wu2021seqformer}       & R50 &Full     & 45.1 & 66.9    & 50.5    & 45.6   & 54.6    \\
SeqFormer~\cite{wu2021seqformer}       & R50 &Full+C80k     & \textbf{47.4} & 69.8    & 51.8    & 45.5   & 54.8    \\
Mask2Former-VIS~\cite{cheng2021mask2former} & R50 &Full     & 46.4 & 68.0    & 50.0    & --     & --      \\
\ours            & R50 &Full     & \textbf{47.4} & 69.0    & \textbf{52.1}    & \textbf{45.7}   & \textbf{55.7}    \\\midrule
TeViT~\cite{yang2022tevit}           & Swin-L &Full  & 56.8 & 80.6    & 63.1    & 52.0   & 63.3    \\
SeqFormer~\cite{wu2021seqformer}       & Swin-L &Full+C80k   & 59.3 & 82.1    & 66.4    & 51.7   & 64.4    \\
Mask2Former-VIS~\cite{cheng2021mask2former} & Swin-L &Full   & 60.4 & \textbf{84.4}    & 67.0    & --     & --      \\
\ours            & Swin-L &Full  & \textbf{61.6} & 83.3    & \textbf{68.6}    & \textbf{54.8}   & \textbf{66.6}    \\\midrule
\ours      & Swin-L &1\%    & 59.0 & 81.6    & 64.7    & 54.0   & 64.0    \\
\ours      & Swin-L &5\%   & 59.3 & 81.4    & 65.8    & 53.8   & 64.1    \\
\ours    & Swin-L &10\%   & 61.0 & 83.0    & 67.7    & 54.6   & 66.1    \\
    \bottomrule
  \end{tabular}
\end{table}

\noindent\textbf{YouTube-VIS 2019.} The results for YouTube-VIS 2019 are shown in \tabref{ytvis2019}. 
\ours achieves highest AP and most other metrics for both ResNet-50 and Swin-L backbones.  SeqFormer shows that it is beneficial to jointly train with  images from COCO~\cite{lin2014microsoft} that contain YouTubeVIS categories (+C80k in table). TeViT proposes messenger shift transformer (MsgShifT) that are as efficient as ResNet backbones, while improving the VIS performances. Our ResNet-50 results match or outperform their results without further modifications. Compared to the state-of-the-art Mask2Former-VIS, which can be seen as the per-clip approach to apply Mask2Former to VIS, \ours consistently outperforms by around 1\% for both backbones.
\ours with X\% means sub-sampling the annotated frames for each video in training.  Since there are less temporal variations for videos in  YouTube-VIS 2019, \ours with 1\% of training frames only reduces AP by 2.6\%. This significantly reduces the annotation effort while not sacrificing much performance.

\begin{table}
  \caption{YouTube-VIS 2021 Results. \ours's performance improvement increases on the more challenging YouTube-VIS 2021. Our 1\% results already outperform previous state-of-the-art.}
  \label{tab:ytvis2021}
  \centering
  \tabfontsize
  \begin{tabular}{lllccccc}
    \toprule
Method          & Backbone & Training  & AP   & AP$_{50}$ & AP$_{75}$ & AR$_1$ & AR$_{10}$ \\\midrule
TeViT~\cite{yang2022tevit}           & MsgShifT & Full      & 37.9 & 61.2    & 42.1    & 35.1   & 44.6    \\
SeqFormer~\cite{wu2021seqformer}       & R50      & Full+C80k & 40.5 & 62.4    & 43.7    & 36.1   & 48.1    \\
Mask2Former-VIS~\cite{cheng2021mask2former} & R50      & Full      & 40.6 & 60.9    & 41.8    & --     & --      \\
\ours            & R50      & Full      & \textbf{44.2} & \textbf{66.0}    & \textbf{48.1}    & \textbf{39.2}   & \textbf{51.7}    \\\midrule
SeqFormer~\cite{wu2021seqformer}       & Swin-L   & Full+C80k & 51.8 & 74.6    & 58.2    & 42.8   & 58.1    \\
Mask2Former-VIS~\cite{cheng2021mask2former} & Swin-L   & Full      & 52.6 & 76.4    & 57.2    & --     & --      \\
\ours            & Swin-L   & Full      & \textbf{55.3} & \textbf{76.6}    & \textbf{62.0}    & \textbf{45.9}   & \textbf{60.8}    \\\midrule
\ours            & Swin-L   & 1\%       & 52.9 & 74.9    & 58.9    & 44.7   & 58.3    \\
\ours            & Swin-L   & 5\%       & 54.3 & 76.3    & 60.1    & 45.4   & 59.5    \\
\ours            & Swin-L   & 10\%      & 54.9 & 76.3    & 61.9    & 45.3   & 60.1    \\
    \bottomrule
  \end{tabular}
\end{table}

\noindent\textbf{YouTube-VIS 2021.} The results for YouTube-VIS 2021 are shown in \tabref{ytvis2021}. On this more challenging dataset, the performance improvements for \ours increase compared to YouTube-VIS 2019. Without better backbone like TeViT and additional training data like SeqFormer, our ResNet-50 results outperform by a large margin for all metrics. This is the also case for Swin-L. \ours outperforms previous state-of-the-art Mask2Former-VIS by 2.7\%. By using only 1\% of training frames, \ours's AP decrease by only 2.4\%, which means that our 1\% result still outperforms previous state-of-the art. We also see that on YouTube-VIS datasets, reducing the annotations by 10x does not significantly affect the performances (-0.6\% AP for 2019 and -0.4\% AP for 2021).

\begin{table}
  \caption{OVIS Results. \ours significantly outperform existing approaches on OVIS. Our image-based framework leads to easier and better learning on this dataset with heavy occlusions.}
  \label{tab:ovis}
  \centering
  \tabfontsize
  \begin{tabular}{lllccccc}
    \toprule
Method               & Backbone & Training & AP   & AP$_{50}$ & AP$_{75}$ & AR$_1$ & AR$_{10}$ \\\midrule
TeViT~\cite{yang2022tevit}                & MsgShifT & Full     & 17.4 & 34.9      & 15.0      & 11.2   & 21.8      \\
CrossVIS~\cite{yang2021crossover}             & R50      & Full     & 14.9 & 32.7      & 12.1      & 10.3   & 19.8      \\
CMaskTrack R-CNN~\cite{qi2021occluded}     & R50      & Full     & 15.4 & 33.9      & 13.1      & 9.3    & 20.0      \\
STC~\cite{jiang2022stc}                  & R50      & Full     & 15.5 & 33.5      & 13.4      & 11.0   & 20.8      \\
Mask2Former-VIS*     & R50      & Full     & 17.3 & 37.3      & 15.1      & 10.5   & 23.5      \\
\ours                 & R50      & Full     & \textbf{25.0} & \textbf{45.5}      & \textbf{24.0}      & \textbf{13.9}   & \textbf{29.7}      \\\midrule
MaskTrack R-CNN*+SWA~\cite{li2021limited} & Swin-L   & Full     & 28.9 & 56.3      & 26.8      & 13.5   & 34.0      \\
Mask2Former-VIS*    & Swin-L   & Full     & 25.8 & 46.5      & 24.4      & 13.7   & 32.2      \\
\ours                 & Swin-L   & Full     & \textbf{39.4} & \textbf{61.5}      & \textbf{41.3}      & \textbf{18.1}   & \textbf{43.3}      \\\midrule
\ours                 & Swin-L   & 1\%      & 31.7 & 54.9      & 31.3      & 16.3   & 36.1      \\
\ours                 & Swin-L   & 5\%      & 35.7 & 60.1      & 35.8      & 17.3   & 39.9      \\
\ours                 & Swin-L   & 10\%     & 37.2 & 60.7      & 38.0      & 17.3   & 41.1      \\
    \bottomrule
  \end{tabular}
\end{table}

\begin{figure}
  \centering
  \includegraphics[width=1.0\linewidth]{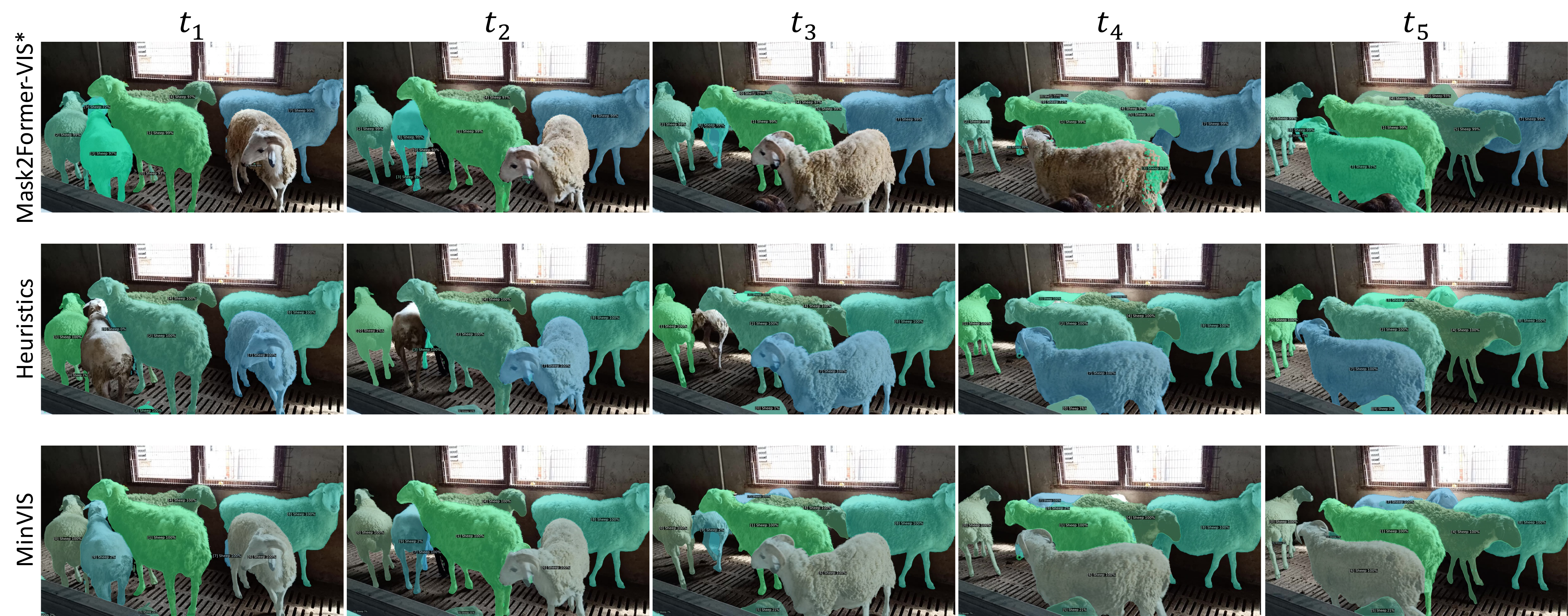}
  \caption{Qualitative results on OVIS. \ours stably tracks all the sheep in the video. 
  Using mask overlap based heuristics instead leads to multiple identity switches in tracking.
  Mask2Former-VIS* uses per-clip training that is difficult to optimize on the challenging OVIS dataset.}
  \label{fig:qual}
\end{figure}

\noindent\textbf{Occluded VIS (OVIS).}  The results for OVIS are shown in \tabref{ovis}. 
Mask2Former-VIS* denotes our application of Mask2Former-VIS to OVIS. Since videos in OVIS can have up to hundreds of frames, we apply Mask2Former-VIS to non-overlapping sliding windows of length 30. The outputs from these clips are then merged by our post-processing. \ours is an online method and does not need modification to apply to OVIS. 
\ours shows significant improvement compared to existing works on OVIS. 
With ResNet-50 backbone, \ours outperforms previous state-of-the-art TeViT with MsgShifT backbone by 7.6\% AP. With Swin-L backbone, \ours outperforms previous best result MaskTrack R-CNN*+SWA by 10.5\% AP, which is the winner of the 1st OVIS Challenge. Their key observation is that sampling frames that are far apart in OVIS leads to drastically different features and makes it hard to train MaskTrack R-CNN. This is in contrast to YouTube-VIS datasets, in which the videos are shorter and there are less temporal variations within the video. We observe the same phenomenon when training Mask2Former-VIS*. However, the limited sampling reference frame strategy of MaskTrack R-CNN*+SWA still does not work in this case. Mask2Former-VIS* uses a fully end-to-end loss instead of an explicit tracking loss to learn temporal association, which makes the learning even harder in OVIS. On the other hand, \ours is image-based and does not need to worry about the temporal sampling strategy to train the model. This is contrary to common belief that per-frame approaches are worse for scenarios with heavy occlusions. Instead, our image-based approach leads to easier and better learning on OVIS. 
We show that an image instance segmentation model that can segment occluded instances in each frame is also good at associating such instances across frames. 
\figref{qual} shows qualitative results. \ours stably tracks all the sheep in the video. Using mask overlap based heuristics instead of query matching leads to multiple identity switches in tracking.
Mask2Former-VIS* does not have as good segmentation masks because its training is interfered by heavy occlusions and large appearance deformations between frames in OVIS.  

\begin{figure}
  \centering
  \includegraphics[width=1.0\linewidth]{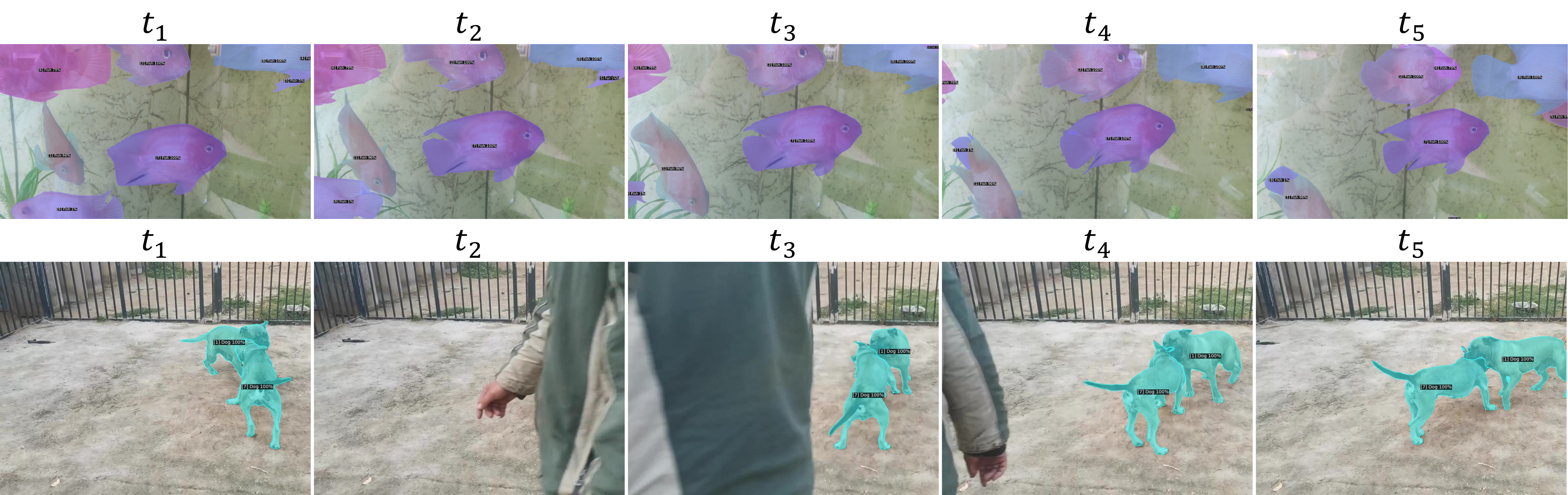}
  \caption{Failure cases of \ours on OVIS. When an object instance disappear from a video, \ours can fail by associating its query embedding to a wrong mask without overlap (top). This is because we do not use mask overlap heuristics in our work. On the other hand, we are also limited by the image instance segmentation model, which might not work well on close-up objects (bottom). }
  \label{fig:qual-fail}
\end{figure}

\figref{qual-fail} shows additional qualitative results on failure cases of \ours on the OVIS dataset. As discussed in \secref{match}, \ours does not use heuristics to handle the birth and death of object instances. The death of an object instance is correctly handled if its query is matched to a query in the next frame that produces an empty mask. Despite its simplicity and effectiveness, the drawback of this approach is that there is nothing stopping the model from matching the disappearing query to a query with a non-empty mask. From $t_3$ to $t_4$ in the top row of \figref{qual-fail}, as the fish in the lower left leaves the frame, \ours associates it to a mask covering the tail of a nearby fish. From $t_4$ to $t_5$, when the fish in the upper left leaves the frame, \ours again associates it to a mask covering the head of a nearby fish. Since the associated masks are non-empty, \ours fails to correctly handle the death of these instances. On the other hand, when the two dogs in the bottom row of \figref{qual-fail} are covered in $t_2$, \ours correctly associates their queries to empty masks. \ours further correctly handles the object births in $t_3$. However, \ours is limited by the segmentation of the image segmentation model, which fails to segment the close-up person.

\subsection{Analyzing Query Matching}
\label{sec:analysis}

\begin{table}
  \caption{Comparison of post-processing. Heuristics based on mask over laps lead to significant AP drop on OVIS. Our query matching approach has simpler design without loss of performance.}
  \label{tab:heu}
  \centering
  \tabfontsize
  \begin{tabular}{llccccc}
    \toprule
Method             & Dataset          & AP   & AP$_{50}$ & AP$_{75}$ & AR$_1$ & AR$_{10}$ \\\midrule
heuristics only    & YouTube-VIS 2019 & 58.2 & 79.2      & 64.1      & 51.3   & 63.6      \\
heuristics + query & YouTube-VIS 2019 & 61.3 & 82.8      & \textbf{68.7}      & 54.3   & 66.3      \\
query only         & YouTube-VIS 2019 & \textbf{61.6} & \textbf{83.3}      & 68.6      & \textbf{54.8}   & \textbf{66.6}      \\\midrule
heuristics only    & YouTube-VIS 2021 & 52.7 & 75.3      & 57.3      & 44.4   & 58.3      \\
heuristics + query & YouTube-VIS 2021 & 55.1 & 76.2      & 61.9      & \textbf{46.0}   & 60.7      \\
query only         & YouTube-VIS 2021 & \textbf{55.3} & \textbf{76.6}      & \textbf{62.0}      & 45.9   & \textbf{60.8}      \\\midrule
heuristics only    & Occluded VIS     & 31.7 & 56.0      & 31.3      & 15.8   & 35.8      \\
heuristics + query & Occluded VIS     & 39.1 & \textbf{62.5}      & 40.8      & 17.7   & \textbf{43.4}      \\
query only         & Occluded VIS     & \textbf{39.4} & 61.5      & \textbf{41.3}      & \textbf{18.1}   & 43.3      \\
    \bottomrule
  \end{tabular}
\end{table}

The success of \ours depends on whether query matching is good for tracking instances. We conduct ablation studies by comparing it to manually designed heuristics. We use the bipartite matching heuristics in \secref{training} for tracking by treating instances in the last frame as groundtruth. The results are in \tabref{heu}. Using heuristics lead to around 3\% AP drop on both YouTube-VIS 2019 and 2021. It leads to more significant drop on OVIS (7.7\%) due to heavier occlusions. We also combine query matching and heuristics with equal weights, which has mixed results. Our query only approach is simpler and more generalizable without loss of performance.

We visualize the learned query embeddings by t-SNE~\cite{van2008visualizing} in \figref{tsne}. Each plot is for a video in the training set. We visualize the training set to see the effect of an image only objective (to segment instances in an image) on query embeddings across different frames. Query embeddings of the same instance have the same color. We obtain the instance IDs for queries by bipartite matching its outputs to groundtruth instances, which have consistent IDs across frames. Without any video-based tracking objective, query embeddings of the same instances are already grouped into distinct clusters, even for the OVIS dataset. This supports our design of only using image-based objectives. In Appendix~\ref{sec:tsne-val}, we further visualize query embeddings on videos not used in training.

\begin{table}
  \caption{Results for adding supervision to query matching. The supervision can provide dataset dependent benefit if the temporal hyper-parameters are selected properly. }
  \label{tab:sup}
  \centering
  \tabfontsize
  \begin{tabular}{llccccc}
    \toprule
Method                & Dataset          & AP   & AP$_{50}$ & AP$_{75}$ & AR$_1$ & AR$_{10}$ \\\midrule
\ours                  & YouTube-VIS 2019 & \textbf{61.6} & \textbf{83.3}      & \textbf{68.6}      & \textbf{54.8}   & \textbf{66.6}      \\
+ Supervised Matching & YouTube-VIS 2019 & 61.0 & 82.1      & 67.6      & 54.3   & 66.1      \\
+ Limited Range       & YouTube-VIS 2019 & 60.7 & 82.5      & 67.0      & 54.1   & 65.5      \\\midrule
\ours                  & YouTube-VIS 2021 & \textbf{55.3} & 76.6      & \textbf{62.0}      & \textbf{45.9}   & \textbf{60.8}      \\
+ Supervised Matching & YouTube-VIS 2021 & 54.4 & 75.7      & 60.6      & 45.5   & 59.5      \\
+ Limited Range       & YouTube-VIS 2021 & 55.2 & \textbf{77.0}      & 61.5      & 45.4   & 60.1      \\\midrule
\ours                  & Occluded VIS    & 39.4 & 61.5      & \textbf{41.3}      & 18.1   & \textbf{43.3}      \\
+ Supervised Matching & Occluded VIS     & 38.7 & 61.2      & 39.6      & 17.9   & 42.4      \\
+ Limited Range       & Occluded VIS     & \textbf{39.6} & \textbf{63.2}      & 41.0      & \textbf{18.2}   & 43.0      \\
    \bottomrule
  \end{tabular}
\end{table}

\subsection{Effect of Video-based Training} 
\label{sec:sup}

While we have shown that \ours achieves state-of-the-art VIS performance without video-based training, it is interesting to see how we can leverage video annotation when it is available. We use the video annotation to supervise our matching as in previous per-frame works~\cite{yang2019video,yang2021crossover,QueryTrack}. Given two sampled frames, we use a hinge loss  to  ensure that the correct associations of queries have the highest inner products compared to that of other queries between the two frames~\cite{QueryTrack}. The results are in  \tabref{sup}. The ``Supervised Matching'' rows mean directly applying the matching supervision to the original frame sampling process. In our case, this means that the two sampled frames might be separated up to 20 frames. As pointed out in previous works, frames that are far separated increase the training difficulty and can hurt model performances especially with occlusions~\cite{li2021limited}. We thus also consider the ``Limited Range'' training to only sample consecutive frames for supervised matching, as we only need to match consecutive frames. From the results, directly applying ``Supervised Matching'' hurt performances on all three datasets. Adding ``Limited Range'' recovers most of the performances for YouTube-VIS 2021 and OVIS. On OVIS, it even marginally outperforms the original \ours. However, this improvement does rely on the dataset dependent sampling range. We believe it is possible and important to use video-based training to further improve \ours, although this would take away \ours's practical advantages of only needing sparse annotations and having a simple training pipeline. Appendix~\ref{sec:limit} discusses further limitations of not using video information in training.
\section{Conclusion}

We show that a purely image-based training procedure can lead to competitive performances for VIS. Our key finding is that instance tracking naturally emerges in query-based image instance segmentation models with proper architectural constraints. 
In addition to improving state-of-the-art approaches on YouTube-VIS 2019/2021, we show that this is particularly beneficial for OVIS. The image-based objective reduces the learning difficulty and leads to better performances. 
\ours only requires sparse frame annotations, which makes it much more applicable to real-world scenarios.
We believe a promising direction to extend \ours is to explore ways to better leverage the video frames that are not annotated to further improve our performances with sub-sampled annotations.


\bibliographystyle{unsrt}
\bibliography{vis}

\begin{thebibliography}{10}

\bibitem{yang2019video}
Linjie Yang, Yuchen Fan, and Ning Xu.
\newblock Video instance segmentation.
\newblock In {\em ICCV}, 2019.

\bibitem{yang2021crossover}
Shusheng Yang, Yuxin Fang, Xinggang Wang, Yu~Li, Chen Fang, Ying Shan, Bin
  Feng, and Wenyu Liu.
\newblock Crossover learning for fast online video instance segmentation.
\newblock In {\em ICCV}, 2021.

\bibitem{cheng2021mask2former}
Bowen Cheng, Anwesa Choudhuri, Ishan Misra, Alexander Kirillov, Rohit Girdhar,
  and Alexander~G Schwing.
\newblock Mask2former for video instance segmentation.
\newblock {\em arXiv preprint arXiv:2112.10764}, 2021.

\bibitem{wang2021end}
Yuqing Wang, Zhaoliang Xu, Xinlong Wang, Chunhua Shen, Baoshan Cheng, Hao Shen,
  and Huaxia Xia.
\newblock End-to-end video instance segmentation with transformers.
\newblock In {\em CVPR}, 2021.

\bibitem{wu2021seqformer}
Junfeng Wu, Yi~Jiang, Wenqing Zhang, Xiang Bai, and Song Bai.
\newblock Seqformer: a frustratingly simple model for video instance
  segmentation.
\newblock {\em arXiv preprint arXiv:2112.08275}, 2021.

\bibitem{carion2020end}
Nicolas Carion, Francisco Massa, Gabriel Synnaeve, Nicolas Usunier, Alexander
  Kirillov, and Sergey Zagoruyko.
\newblock End-to-end object detection with transformers.
\newblock In {\em ECCV}, 2020.

\bibitem{hwang2021video}
Sukjun Hwang, Miran Heo, Seoung~Wug Oh, and Seon~Joo Kim.
\newblock Video instance segmentation using inter-frame communication
  transformers.
\newblock {\em NeurIPS}, 2021.

\bibitem{yang2022tevit}
Shusheng Yang, Xinggang Wang, Yu~Li, Yuxin Fang, Jiemin Fang, Liu, Xun Zhao,
  and Ying Shan.
\newblock Temporally efficient vision transformer for video instance
  segmentation.
\newblock In {\em CVPR}, 2022.

\bibitem{fu2021learning}
Yang Fu, Sifei Liu, Umar Iqbal, Shalini De~Mello, Humphrey Shi, and Jan Kautz.
\newblock Learning to track instances without video annotations.
\newblock In {\em CVPR}, 2021.

\bibitem{liu2021weakly}
Qing Liu, Vignesh Ramanathan, Dhruv Mahajan, Alan Yuille, and Zhenheng Yang.
\newblock Weakly supervised instance segmentation for videos with temporal mask
  consistency.
\newblock In {\em CVPR}, 2021.

\bibitem{QueryTrack}
Shusheng Yang, Yuxin Fang, Xinggang Wang, Yu~Li, Ying Shan, Bin Feng, and Wenyu
  Liu.
\newblock Tracking instances as queries.
\newblock {\em arXiv preprint arXiv:2106.11963}, 2021.

\bibitem{qi2021occluded}
Jiyang Qi, Yan Gao, Yao Hu, Xinggang Wang, Xiaoyu Liu, Xiang Bai, Serge
  Belongie, Alan Yuille, Philip Torr, and Song Bai.
\newblock Occluded video instance segmentation: A benchmark.
\newblock {\em arXiv preprint arXiv:2102.01558}, 2021.

\bibitem{liu2021swin}
Ze~Liu, Yutong Lin, Yue Cao, Han Hu, Yixuan Wei, Zheng Zhang, Stephen Lin, and
  Baining Guo.
\newblock Swin transformer: Hierarchical vision transformer using shifted
  windows.
\newblock In {\em ICCV}, 2021.

\bibitem{li2021limited}
Zhuang Li, Leilei Cao, and Hongbin Wang.
\newblock Limited sampling reference frame for masktrack r-cnn.
\newblock In {\em ICCVW}, 2021.

\bibitem{he2017mask}
Kaiming He, Georgia Gkioxari, Piotr Doll{\'a}r, and Ross Girshick.
\newblock Mask r-cnn.
\newblock In {\em ICVV}, 2017.

\bibitem{bertasius2020classifying}
Gedas Bertasius and Lorenzo Torresani.
\newblock Classifying, segmenting, and tracking object instances in video with
  mask propagation.
\newblock In {\em CVPR}, 2020.

\bibitem{Fang_2021_ICCV}
Yuxin Fang, Shusheng Yang, Xinggang Wang, Yu~Li, Chen Fang, Ying Shan, Bin
  Feng, and Wenyu Liu.
\newblock Instances as queries.
\newblock In {\em ICCV}, 2021.

\bibitem{IDOL}
Junfeng Wu, Qihao Liu, Yi~Jiang, Song Bai, Alan Yuille, and Xiang Bai.
\newblock In defense of online models for video instance segmentation.
\newblock In {\em ECCV}, 2022.

\bibitem{zeng2021motr}
Fangao Zeng, Bin Dong, Yuang Zhang, Tiancai Wang, Xiangyu Zhang, and Yichen
  Wei.
\newblock Motr: End-to-end multiple-object tracking with transformer.
\newblock In {\em ECCV}, 2022.

\bibitem{meinhardt2021trackformer}
Tim Meinhardt, Alexander Kirillov, Laura Leal-Taixe, and Christoph
  Feichtenhofer.
\newblock Trackformer: Multi-object tracking with transformers.
\newblock In {\em CVPR}, 2022.

\bibitem{lu2020learning}
Xiankai Lu, Wenguan Wang, Jianbing Shen, Yu-Wing Tai, David~J Crandall, and
  Steven~CH Hoi.
\newblock Learning video object segmentation from unlabeled videos.
\newblock In {\em CVPR}, 2020.

\bibitem{voigtlaender2021reducing}
Paul Voigtlaender, Lishu Luo, Chun Yuan, Yong Jiang, and Bastian Leibe.
\newblock Reducing the annotation effort for video object segmentation
  datasets.
\newblock In {\em WACV}, 2021.

\bibitem{yang2021dystab}
Yanchao Yang, Brian Lai, and Stefano Soatto.
\newblock Dystab: Unsupervised object segmentation via dynamic-static
  bootstrapping.
\newblock In {\em CVPR}, 2021.

\bibitem{ahn2019weakly}
Jiwoon Ahn, Sunghyun Cho, and Suha Kwak.
\newblock Weakly supervised learning of instance segmentation with inter-pixel
  relations.
\newblock In {\em CVPR}, 2019.

\bibitem{lan2021discobox}
Shiyi Lan, Zhiding Yu, Christopher Choy, Subhashree Radhakrishnan, Guilin Liu,
  Yuke Zhu, Larry~S Davis, and Anima Anandkumar.
\newblock Discobox: Weakly supervised instance segmentation and semantic
  correspondence from box supervision.
\newblock In {\em ICCV}, 2021.

\bibitem{tian2021boxinst}
Zhi Tian, Chunhua Shen, Xinlong Wang, and Hao Chen.
\newblock Boxinst: High-performance instance segmentation with box annotations.
\newblock In {\em CVPR}, 2021.

\bibitem{wang2022freesolo}
Xinlong Wang, Zhiding Yu, Shalini De~Mello, Jan Kautz, Anima Anandkumar,
  Chunhua Shen, and Jose~M Alvarez.
\newblock Freesolo: Learning to segment objects without annotations.
\newblock In {\em CVPR}, 2022.

\bibitem{wang2021survey}
Wenguan Wang, Tianfei Zhou, Fatih Porikli, David Crandall, and Luc Van~Gool.
\newblock A survey on deep learning technique for video segmentation.
\newblock {\em arXiv preprint arXiv:2107.01153}, 2021.

\bibitem{wang2020solo}
Xinlong Wang, Tao Kong, Chunhua Shen, Yuning Jiang, and Lei Li.
\newblock Solo: Segmenting objects by locations.
\newblock In {\em ECCV}, 2020.

\bibitem{cheng2021masked}
Bowen Cheng, Ishan Misra, Alexander~G Schwing, Alexander Kirillov, and Rohit
  Girdhar.
\newblock Masked-attention mask transformer for universal image segmentation.
\newblock In {\em CVPR}, 2022.

\bibitem{zhu2021deformable}
Xizhou Zhu, Weijie Su, Lewei Lu, Bin Li, Xiaogang Wang, and Jifeng Dai.
\newblock Deformable detr: Deformable transformers for end-to-end object
  detection.
\newblock In {\em ICLR}, 2021.

\bibitem{milletari2016v}
Fausto Milletari, Nassir Navab, and Seyed-Ahmad Ahmadi.
\newblock V-net: Fully convolutional neural networks for volumetric medical
  image segmentation.
\newblock In {\em 3DV}, 2016.

\bibitem{jiang2022stc}
Zhengkai Jiang, Zhangxuan Gu, Jinlong Peng, Hang Zhou, Liang Liu, Yabiao Wang,
  Ying Tai, Chengjie Wang, and Liqing Zhang.
\newblock Stc: Spatio-temporal contrastive learning for video instance
  segmentation.
\newblock {\em arXiv preprint arXiv:2202.03747}, 2022.

\bibitem{athar2020stem}
Ali Athar, Sabarinath Mahadevan, Aljosa Osep, Laura Leal-Taix{\'e}, and Bastian
  Leibe.
\newblock Stem-seg: Spatio-temporal embeddings for instance segmentation in
  videos.
\newblock In {\em ECCV}, 2020.

\bibitem{lin2014microsoft}
Tsung-Yi Lin, Michael Maire, Serge Belongie, James Hays, Pietro Perona, Deva
  Ramanan, Piotr Doll{\'a}r, and C~Lawrence Zitnick.
\newblock Microsoft coco: Common objects in context.
\newblock In {\em ECCV}, 2014.

\bibitem{van2008visualizing}
Laurens Van~der Maaten and Geoffrey Hinton.
\newblock Visualizing data using t-sne.
\newblock {\em JMLR}, 9(11), 2008.

\bibitem{xu2018youtube}
Ning Xu, Linjie Yang, Yuchen Fan, Dingcheng Yue, Yuchen Liang, Jianchao Yang,
  and Thomas Huang.
\newblock Youtube-vos: A large-scale video object segmentation benchmark.
\newblock {\em arXiv preprint arXiv:1809.03327}, 2018.

\bibitem{abu2016youtube}
Sami Abu-El-Haija, Nisarg Kothari, Joonseok Lee, Paul Natsev, George Toderici,
  Balakrishnan Varadarajan, and Sudheendra Vijayanarasimhan.
\newblock Youtube-8m: A large-scale video classification benchmark.
\newblock {\em arXiv preprint arXiv:1609.08675}, 2016.

\end{thebibliography}

\appendix
\section{Limitations and Potential Negative Social Impacts}
\label{sec:limit}

\noindent\textbf{Limitation.} We have discussed in the main paper on the possibility of improving \ours with video-based training. While we believe there are practical advantages of using our image-based VIS training pipeline, videos provide lots of extra information that we are not currently leveraging. In particular, temporal supervision from video should make our query embeddings even more suitable for tracking instances. In \figref{qual-fail}, we visualize failure cases of using our current query embeddings for tracking. We conduct further analysis of Supervised Matching in Appendix~\ref{sec:sqm} and believe further investigation along this direction should improve our approach.
In addition to improving fully-supervised performance, we believe a promising direction is to explore semi-supervised learning at the frame level. In this case, one can temporally propagate the sub-sampled annotations in training to further improve the performance with reduced supervision.

\noindent\textbf{Potential Negative Social Impacts.} Video instance segmentation is a challenging video task, and thus provides fine-grained understanding of videos. The tracking and segmentation of objects of interest might be use for surveillance applications with negative social impact. While ``person'' is a category in the datasets used in this paper, no further protected attributes are annotated. Therefore, our trained models' performance on human subjects might not be fair with respect to protected attributes.

\section{Further Details for Datasets}

The YouTube-VIS 2019/2021 datasets are under \href{https://creativecommons.org/licenses/by/4.0/}{CC BY 4.0 License}, and Occluded VIS is under \href{https://creativecommons.org/licenses/by-nc-sa/4.0/}{CC BY-NC-SA 4.0 License}. The videos in YouTube-VIS are from YouTube-VOS~\cite{xu2018youtube}, whose videos are in turn from YouTube-8M~\cite{abu2016youtube}. YouTube-8M uses public videos on YouTube but does not discuss the process to filter personally identifiable information or offensive content in the paper.

\begin{table}[t]
  \caption{YouTube-VIS 2019 results. C80k indicates joint training with COCO images that have YouTube-VIS categories. \ours with X\% means sub-sampling the annotated frames in training.}
  \label{tab:ytvis2019_std}
  \centering
  \scriptsize
  \begin{tabular}{lllccccc}
    \toprule
Method         & Backbone & Training  & AP   & AP$_{50}$ & AP$_{75}$ & AR$_1$ & AR$_{10}$ \\\midrule
TeViT~\cite{yang2022tevit}           & R50  &Full      & 42.1 & 67.8    & 44.8    & 41.3   & 49.4    \\
TeViT~\cite{yang2022tevit}           & MsgShifT  &Full      & 46.6 & \textbf{71.3}    & 51.6    & 44.9   & 54.3    \\
SeqFormer~\cite{wu2021seqformer}       & R50 &Full     & 45.1 & 66.9    & 50.5    & 45.6   & 54.6    \\
SeqFormer~\cite{wu2021seqformer}       & R50 &+C80k     & \textbf{47.4} & 69.8    & 51.8    & 45.5   & 54.8    \\
Mask2Former~\cite{cheng2021mask2former} & R50 &Full     & 46.4 & 68.0    & 50.0    & --     & --      \\
\ours            & R50 &Full     & \textbf{47.4} $\pm$0.2 & 69.0 $\pm$2.1   & \textbf{52.1} $\pm$0.2   & \textbf{45.7}$\pm$0.2   & \textbf{55.7} $\pm$0.7   \\\midrule
TeViT~\cite{yang2022tevit}           & Swin-L &Full  & 56.8 & 80.6    & 63.1    & 52.0   & 63.3    \\
SeqFormer~\cite{wu2021seqformer}       & Swin-L &+C80k   & 59.3 & 82.1    & 66.4    & 51.7   & 64.4    \\
Mask2Former~\cite{cheng2021mask2former} & Swin-L &Full   & 60.4 & \textbf{84.4}    & 67.0    & --     & --      \\
\ours            & Swin-L &Full  & \textbf{61.6}$\pm$0.3 & 83.3$\pm$0.2    & \textbf{68.6}$\pm$1.6    & \textbf{54.8}$\pm$0.4   & \textbf{66.6} $\pm$0.9   \\\midrule
\ours      & Swin-L &1\%    & 59.0$\pm$0.3 & 81.6 $\pm$0.4   & 64.7$\pm$1.3    & 54.0 $\pm$0.3  & 64.0 $\pm$0.4  \\
\ours      & Swin-L &5\%   & 59.3$\pm$0.2 & 81.4 $\pm$1.7   & 65.8 $\pm$0.7  & 53.8$\pm$0.4   & 64.1 $\pm$0.2   \\
\ours    & Swin-L &10\%   & 61.0 $\pm$0.7 & 83.0$\pm$0.8    & 67.7 $\pm$1.8   & 54.6$\pm$0.3   & 66.1$\pm$0.1    \\
    \bottomrule
  \end{tabular}
\end{table}

\begin{table}[t]
  \caption{YouTube-VIS 2021 Results. \ours's performance improvement increases on the more challenging YouTube-VIS 2021. Our 1\% results already outperform previous state-of-the-art.}
  \label{tab:ytvis2021_std}
  \centering
  \scriptsize
  \begin{tabular}{lllccccc}
    \toprule
Method          & Backbone & Training  & AP   & AP$_{50}$ & AP$_{75}$ & AR$_1$ & AR$_{10}$ \\\midrule
TeViT~\cite{yang2022tevit}           & MsgShifT & Full      & 37.9 & 61.2    & 42.1    & 35.1   & 44.6    \\
SeqFormer~\cite{wu2021seqformer}       & R50      & +C80k & 40.5 & 62.4    & 43.7    & 36.1   & 48.1    \\
Mask2Former~\cite{cheng2021mask2former} & R50      & Full      & 40.6 & 60.9    & 41.8    & --     & --      \\
\ours            & R50      & Full      & \textbf{44.2}$\pm$0.3 & \textbf{66.0}$\pm$0.1    & \textbf{48.1}$\pm$0.7    & \textbf{39.2} $\pm$0.3  & \textbf{51.7} $\pm$0.7   \\\midrule
SeqFormer~\cite{wu2021seqformer}       & Swin-L   & +C80k & 51.8 & 74.6    & 58.2    & 42.8   & 58.1    \\
Mask2Former~\cite{cheng2021mask2former} & Swin-L   & Full      & 52.6 & 76.4    & 57.2    & --     & --      \\
\ours            & Swin-L   & Full      & \textbf{55.3}$\pm$0.2 & \textbf{76.6}$\pm$0.3    & \textbf{62.0}$\pm$0.8    & \textbf{45.9}$\pm$0.2   & \textbf{60.8} $\pm$0.3   \\\midrule
\ours            & Swin-L   & 1\%       & 52.9 $\pm$0.4 & 74.9$\pm$0.5    & 58.9 $\pm$0.7   & 44.7 $\pm$0.3  & 58.3  $\pm$0.7  \\
\ours            & Swin-L   & 5\%       & 54.3 $\pm$0.3 & 76.3 $\pm$0.5   & 60.1$\pm$0.3    & 45.4 $\pm$0.4   & 59.5 $\pm$0.2    \\
\ours            & Swin-L   & 10\%      & 54.9$\pm$0.3 & 76.3$\pm$0.6    & 61.9$\pm$0.2    & 45.3 $\pm$0.2  & 60.1 $\pm$0.4   \\
    \bottomrule
  \end{tabular}
\end{table}

\begin{table}[t]
  \caption{OVIS Results. \ours significantly outperform existing approaches on OVIS. Our image-based framework leads to easier and better learning on this dataset with heavy occlusions.}
  \label{tab:ovis_std}
  \centering
  \scriptsize
  \begin{tabular}{lllccccc}
    \toprule
Method               & Backbone & Training & AP   & AP$_{50}$ & AP$_{75}$ & AR$_1$ & AR$_{10}$ \\\midrule
TeViT~\cite{yang2022tevit}                & MsgShifT & Full     & 17.4 & 34.9      & 15.0      & 11.2   & 21.8      \\
CrossVIS~\cite{yang2021crossover}             & R50      & Full     & 14.9 & 32.7      & 12.1      & 10.3   & 19.8      \\
CMaskTrack R-CNN~\cite{qi2021occluded}     & R50      & Full     & 15.4 & 33.9      & 13.1      & 9.3    & 20.0      \\
STC~\cite{jiang2022stc}                  & R50      & Full     & 15.5 & 33.5      & 13.4      & 11.0   & 20.8      \\
Mask2Former-VIS*     & R50      & Full     & 17.3 & 37.3      & 15.1      & 10.5   & 23.5      \\
\ours                 & R50      & Full     & \textbf{25.0}$\pm$0.3 & \textbf{45.5}$\pm$0.6      & \textbf{24.0}$\pm$0.7      & \textbf{13.9}$\pm$0.3   & \textbf{29.7}$\pm$0.3      \\\midrule
MaskTrack R-CNN*+SWA~\cite{li2021limited} & Swin-L   & Full     & 28.9 & 56.3      & 26.8      & 13.5   & 34.0      \\
Mask2Former-VIS*    & Swin-L   & Full     & 25.8 & 46.5      & 24.4      & 13.7   & 32.2      \\
\ours                 & Swin-L   & Full     & \textbf{39.4}$\pm$0.5 & \textbf{61.5}$\pm$0.1      & \textbf{41.3}$\pm$0.6      & \textbf{18.1}$\pm$0.1   & \textbf{43.3}$\pm$0.5      \\\midrule
\ours                 & Swin-L   & 1\%      & 31.7$\pm$0.5 & 54.9 $\pm$1.0     & 31.3$\pm$0.5      & 16.3$\pm$0.3   & 36.1$\pm$0.3     \\
\ours                 & Swin-L   & 5\%      & 35.7$\pm$0.4 & 60.1 $\pm$1.2     & 35.8$\pm$0.7      & 17.3$\pm$0.1   & 39.9$\pm$0.3      \\
\ours                 & Swin-L   & 10\%     & 37.2$\pm$0.5 & 60.7 $\pm$1.1     & 38.0$\pm$1.0      & 17.3$\pm$0.2   & 41.1$\pm$0.4      \\
    \bottomrule
  \end{tabular}
\end{table}

\section{Tables with Standard Deviation}

Tables with standard deviations are shown in \tabref{ytvis2019_std}, \tabref{ytvis2021_std}, and \tabref{ovis_std}.

\section{Reducing Supervision for Mask2Former-VIS}

\begin{table}
  \caption{Sub-sampling the annotated training frames for \ours and Mask2Former-VIS. \ours outperforms Mask2Former-VIS for all of our settings. The improvement of \ours increases as we further sub-sample the annotated frames.}
  \label{tab:subsample}
  \centering
  \tabfontsize
  \begin{tabular}{lllcccc}
    \toprule
Method         & Dataset & Backbone  & Full   & 10\% & 5\% & 1\%  \\\midrule
Mask2Former-VIS & YTVIS-19 &Swin-L   & 60.4 & 59.0    & 57.8    & 57.3          \\
\ours            & YTVIS-19 &Swin-L  & \textbf{61.6} & \textbf{61.0}    & \textbf{59.3}    & \textbf{59.0}      \\\midrule
Mask2Former-VIS      & YTVIS-21 &Swin-L    & 52.6 & 51.2    & 50.0    & 47.1       \\
\ours      & YTVIS-21 &Swin-L    & \textbf{55.3} & \textbf{54.9}   & \textbf{54.3}    & \textbf{52.9}     \\\midrule
Mask2Former-VIS      & OVIS &Swin-L   & 25.8 & 24.1    & 22.3    & 14.5      \\
\ours      & OVIS &Swin-L   & \textbf{39.4} & \textbf{37.2}    & \textbf{35.7}    & \textbf{31.7}      \\
    \bottomrule
  \end{tabular}
\end{table}

The results for sub-sampling annotated frames for Mask2Former-VIS~\cite{cheng2021mask2former} are shown in \tabref{subsample}. \ours consistently outperforms Mask2Former-VIS in all settings. The improvement increases for all three datasets when we  sub-sample the annotation: +1.2\% for full supervision v.s. +1.7\% for 1\% supervision on YouTube-VIS 2019. +2.7\% for full supervision v.s. +5.8\% for 1\% supervision on YouTube-VIS 2021. +13.6\% for full supervision v.s. +17.2\% for 1\% supervision on OVIS.

\section{Visualizing Query Embeddings in Evaluation}
\label{sec:tsne-val}

\begin{figure}
  \centering
  \includegraphics[width=1.0\linewidth]{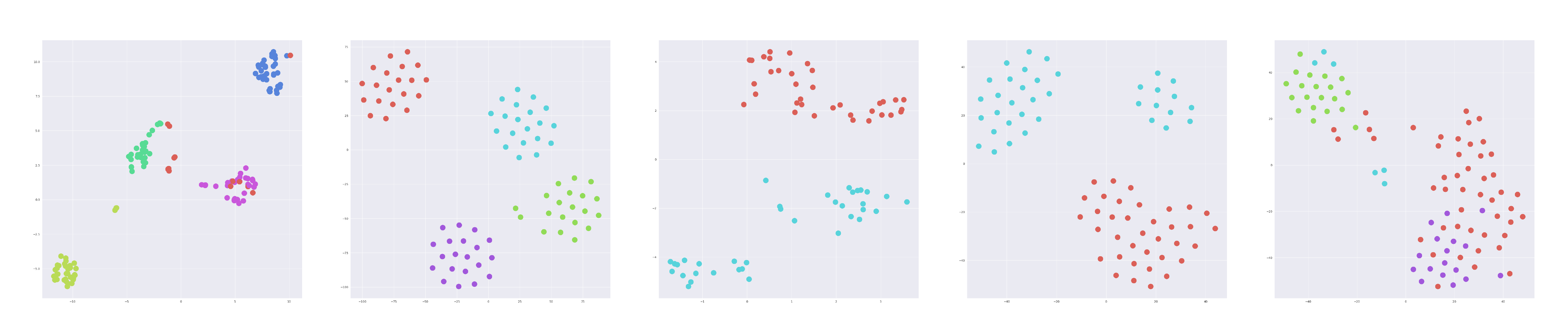}
  \caption{Visualizing our query embeddings during evaluation on videos not used in training. Each plot is for a video, and query embeddings of the same instance (from different frames) have the same color. Despite being noisier than training videos, the query embeddings are still grouped into clusters by instance without any video-based training.}
  \label{fig:tsne-val}
\end{figure}

In the main paper, we visualize the learned query embeddings by t-SNE~\cite{van2008visualizing} in \figref{tsne}. The videos in \figref{tsne} are in the training set and the figure is meant to understand how  query embeddings in training cluster by instances without video-based loss function. We can similarly apply the same visualization to videos that are not used in training. One complication here is that this visualization uses groundtruth instance annotations to determine the corresponding instance ID for each query. However, the groundtruth annotation is not publicly available for the three datasets considered in this work. Our reported results are obtained by submitting our predictions to the datasets' evaluation servers. We therefore perform this analysis by training a new model that only uses 90\% of the training videos in YouTube-VIS 2019, and visualize the learned model's query embeddings during evaluation on the 10\% videos that are not used to train the model. While the 10\% videos are not used in training the model, we still have their groundtruth instances for visualization purposes. This provides a realistic approximation of how our query embeddings would look like for videos not used in training. The visualization is in \figref{tsne-val}. Despite being noisier than training videos, the query embeddings are still grouped into clusters by object instances without any video-based training. This is also quantitatively supported by our state-of-the-art VIS performance on the three datasets.

\section{Further Analysis of Supervised Matching}
\label{sec:sqm}

\begin{figure}
  \centering
  \includegraphics[width=1.0\linewidth]{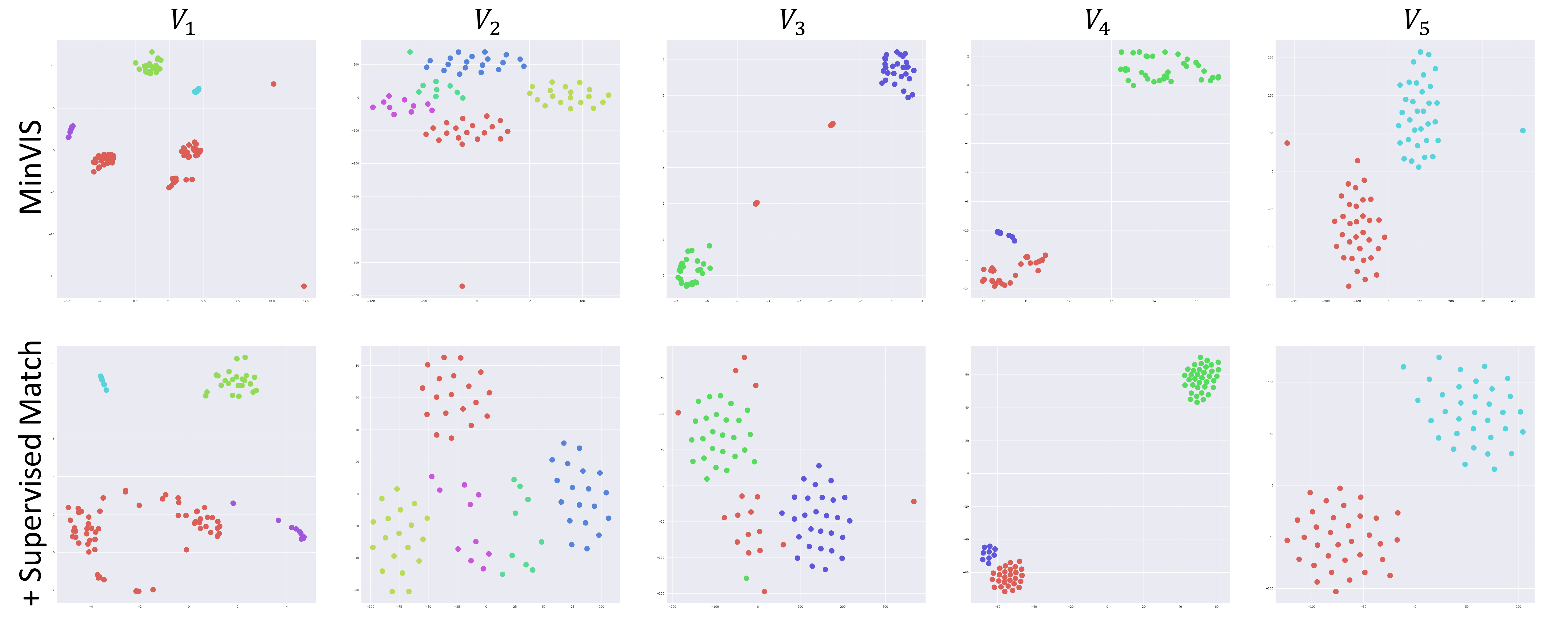}
  \caption{
  Visualizing learned query embeddings on the same videos with and without Supervised Matching. Plots in the same column visualize the same video $V_i$. Supervised matching makes the embeddings more evenly distributed and smooths out the outliers in the embedding space. However, it is unclear whether this is overall beneficial to our tracking by query matching. 
  }
  \label{fig:tsne-sqm}
\end{figure}

We conduct further analysis on the results in \secref{sup}. We visualize the query embeddings on the same training videos with and without using supervised matching. In particular, we perform the analysis on YouTube-VIS 2019 and compare \ours v.s. \ours + Supervised Matching + Limited Range, which hurts performance the most in \tabref{sup}. The visualizations are in \figref{tsne-sqm}. While the plots look similar for most videos, one consistent trend we observe is that adding supervised matching makes the embeddings more evenly distributed and smooths out the outliers in the embedding space. This is a reasonable consequence as the objective encourages the embeddings from the same object instance to be closer to each other. However, it is unclear whether this is overall beneficial to our tracking by query matching. For example, in $V_3$, the outliers are removed at the cost of mixing embeddings from different instances. 
We believe it is an important future work to further understand how to better leverage video information to improve \ours.

\section{Baseline Training Curves on OVIS}

\begin{figure}
  \centering
  \includegraphics[width=1.0\linewidth]{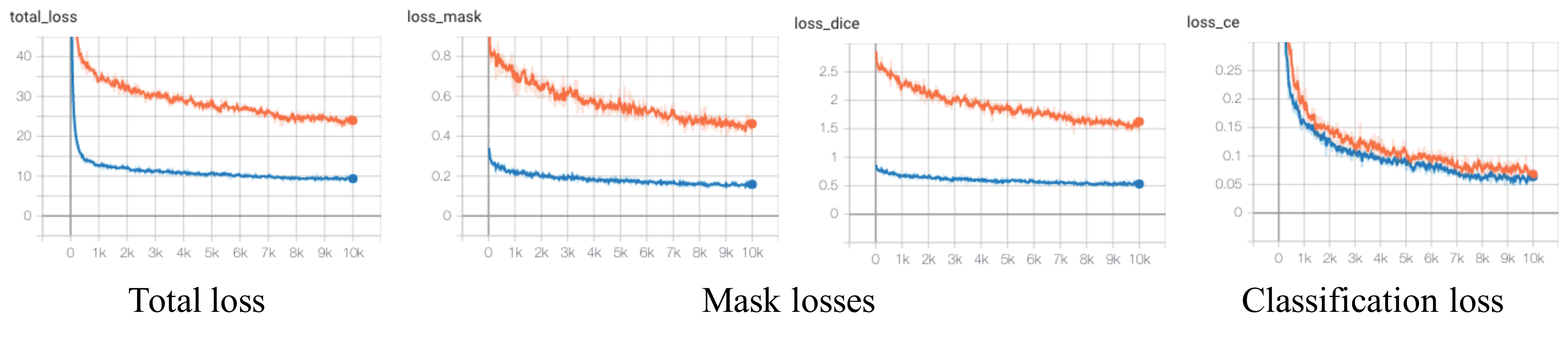}
  \caption{
  Comparing the training curves of \ours and Mask2Former-VIS on OVIS. Blue curves are \ours and orange curves are Mask2Former-VIS.
  }
  \label{fig:baseline}
\end{figure}

As discussed in the main paper, it is difficult to optimize our per-clip baseline on the challenging OVIS dataset. We have included the training curves in \figref{baseline} for further illustration. Blue curves are \ours and orange curves are Mask2Former-VIS. While the classification loss still optimizes well on OVIS, the per-clip baseline has difficulty optimizing  mask related losses.

\end{document}